\documentclass[10pt,twocolumn,letterpaper]{article}

\usepackage{iccv}              

\usepackage{algorithm}
\usepackage{algorithmicx}
\usepackage{algpseudocode}
\usepackage{booktabs} 
\usepackage{multirow}
\usepackage{tabularx}
\usepackage{xcolor}
\usepackage[accsupp]{axessibility}

\definecolor{iccvblue}{rgb}{0.21,0.49,0.74}
\usepackage[pagebackref,breaklinks,colorlinks,allcolors=iccvblue]{hyperref}

\def\paperID{16033} 
\def\confName{ICCV}
\def\confYear{2025}

\title{Context Guided Transformer Entropy Modeling for Video Compression}

\author{Junlong Tong\textsuperscript{1,2}
\quad
Wei Zhang\textsuperscript{2}
\quad
Yaohui Jin\textsuperscript{1}
\quad
Xiaoyu Shen\textsuperscript{2}\thanks{Corresponding author}\\
\textsuperscript{1}Shanghai Jiao Tong University\\
\textsuperscript{2}Ningbo Key Laboratory of Spatial Intelligence and Digital Derivative, Institute of Digital Twin, EIT\\
}

\begin{document}
\maketitle

\begin{abstract}
Conditional entropy models effectively leverage spatio-temporal contexts to reduce video redundancy.
However, incorporating temporal context often introduces additional model complexity and increases computational cost. In parallel, many existing spatial context models lack explicit modeling the ordering of spatial dependencies, which may limit the availability of relevant context during decoding.
To address these issues, we propose the Context Guided Transformer (CGT) entropy model, which estimates probability mass functions of the current frame conditioned on resampled temporal context and dependency-weighted spatial context.
A temporal context resampler learns predefined latent queries to extract critical temporal information using transformer encoders, reducing downstream computational overhead. 
Meanwhile, a teacher-student network is designed as dependency-weighted spatial context assigner to explicitly model the dependency of spatial context order. 
The teacher generates an attention map to represent token importance and an entropy map to reflect prediction certainty from randomly masked inputs, guiding the student to select the weighted top-k tokens with the highest spatial dependency.
During inference, only the student is used to predict undecoded tokens based on high-dependency context.
Experimental results demonstrate that our CGT model reduces entropy modeling time by approximately 65\% and achieves an 11\% BD-Rate reduction compared to the previous state-of-the-art conditional entropy model. \footnote{The code is available at \href{https://github.com/EIT-NLP/CGT}{\nolinkurl{https://github.com/EIT-NLP/CGT}}.}
\end{abstract}

\section{Introduction}
Video compression is essential for the efficient transmission and storage of digital video content. Conventional standards, such as the MPEG series~\cite{le1991mpeg} and HEVC~\cite{sullivan2012overview}, have achieved remarkable success over the past few decades. However, these codecs rely heavily on handcrafted components and domain-specific priors limits their flexibility.

\begin{figure}[t]
    \centering
    \vspace{-1em}
    \includegraphics[width=\linewidth]{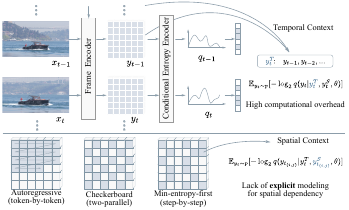}
    \vspace{-2em}
    \caption{ Existing conditional entropy modeling methods incur high computational cost for temporal context modeling, and rely on predefined spatial orders instead of explicit modeling.
    }
    \label{Fig:first}
    \vspace{-1.5em}
\end{figure}

In recent years, deep neural network-based approaches have significantly advanced video compression~\cite{lu2019dvc,habibian2019video,yang2020learning,hu2021fvc,gomes2023video,li2024neural}, achieving substantial improvements in rate-distortion performance.
Among these, conditional entropy modeling \cite{liu2020conditional, mentzer2022vct, li2022hybrid,xiang2023mimt} represents an emerging paradigm, focusing on context-based entropy estimation based on deep neural network.
A key challenge in this framework lies in effectively providing and processing contextual information, including temporal and spatial contexts, to estimate the probability mass functions (PMFs) of video frames.

To exploit temporal context, prior approaches typically employ specialized architectures to capture frame dependencies.
Liu et al. \cite{liu2020conditional} simplified the processing of video frames as a first-order Markov chain to reduce the temporal context to be processed.
Mentzer et al. \cite{mentzer2022vct} employed a Transformer to extract temporal dependencies from two preceding frames using self-attention. 
However, incorporating additional temporal context inevitably increases computational overhead and inference latency, which limits the efficiency of such models in real-time scenarios.

For spatial context, previous methods typically rely on handcrafted or fixed-order decoding strategies.
Autoregressive methods such as VCT~\cite{mentzer2022vct} impose fixed directional assumptions, restricting spatial dependency modeling. Checkerboard decoding~\cite{balle2018variational} enables parallel decoding but still relies on manually predefined context positions. MIMT~\cite{xiang2023mimt} adopts a minimum-entropy principle to iteratively decode with bidirectional context.
By prioritizing regions with the lowest entropy, it reduces bit consumption at each decoding step.
However, these entropy models fail to explicitly model the dependency inherent in spatial context ordering, i.e., identifying the most informative context for undecoded tokens. The limitation hinders the ability to provide the most relevant contextual information during decoding, potentially leading to suboptimal entropy estimation.

To tackle the aforementioned challenges, we propose a context guided transformer (CGT) entropy model, which consists of a temporal context resampler and a dependency-weighted spatial context assigner.
The temporal context resampler leverages a set of learnable window queries as input, performing windowed cross-attention with the temporal context to generate a fixed-length token sequence.
The process utilizes short queries to learn the temporal context for resampling, enabling the learning of temporal dependencies while saving computational cost for subsequent processing.
The dependency-weighted spatial context assigner, built upon a shared-parameter teacher-student swin sransformer network \cite{liu2021swin}, explicitly models spatial position dependencies by balancing token importance (high attention scores) and certainty (low entropy scores). 
The teacher network processes a randomly masked version of the current latent representation, generating an attention map to represent importance and an entropy map to reflect certainty. It then selects the soft top-k \cite{xie2020differentiable, petersen2022differentiable} positions based on a weighted criterion to remove the mask. The student network decodes based on the context obtained after mask removal. 
At inference, only the student network is involved, decoding the weighted top-k positions of the current frame at each step. In short, the dependency-weighted spatial context assigner explicitly models the importance and certainty of context tokens through a teacher-student network and a proxy task based on random masking \cite{devlin2019bert, baobeit}, ensuring consistency between training and inference.

Our contributions can summarized as follows:

\begin{itemize}
    \item 
    We propose a swin transformer-based temporal context resampler that effectively learns temporal dependencies while reducing computational costs. By leveraging learnable queries for resampling, our model generates a compact temporal context, enabling efficient processing while capturing essential temporal information.
    \item 
    We emphasize explicit modeling of spatial context dependencies to distinguish our approach from previous methods. A dependency-weighted spatial context assigner is proposed to balance token importance and certainty. 
    By leveraging a random masking proxy task and a soft top-k selection strategy, the most important tokens are leveraged as contextual guidance for undecoded tokens.
    \item 
    The proposed conditional entropy model is applied to video compression, with extensive experiments demonstrating the effectiveness of our approach.
\end{itemize}
\section{Related Work}

\paragraph{Neural Video Compression}
Recently, deep neural network has greatly promoted the development of video compression.
Lu et al. \cite{lu2019dvc} proposed an innovative end-to-end deep video compression pipeline (DVC), integrating classical video compression architecture with CNNs to jointly optimize motion and residual information for enhanced compression efficiency and video quality.
Going beyond DVC, several studies leverage neural networks to substitute hand-designed modules within the residual coding framework for predictive coding and transformation coding \cite{djelouah2019neural,habibian2019video,hu2020improving,agustsson2020scale}.

Different with the residual coding that follows conventional video codec framework, the DCVC \cite{li2021deep} utilized a deep contextual coding framework for video compression. 
The contextual coding represents a paradigm that focuses on utilizing and learning conditional coding within a deep learning compression framework based on intra-frame and inter-frame context.
To boost the contextual coding, Li et al. \cite{li2023neural} designed a diverse contexts model in both the spatial and temporal dimensions to improve neural video compression performance.
A DCVC-based feature modulation model \cite{li2024neural} is employed to support a wide quality range and maintain effectiveness under long prediction chains.
In addition, Liu et al. \cite{liu2020conditional} proposed a frame-level conditional entropy model aiming to simplify video coding by modeling the correlation between codes for each frame instead of explicitly transforming information across frames as in previous approaches.
Mentzer et al. \cite{mentzer2022vct} proposed a transformer-based conditional entropy model, further advancing the capabilities of conditional models by leveraging transformers to model dependencies more effectively.
Besides, recent methods such as NVRC~\cite{kwan2024nvrc} and MVC~\cite{tang2023scene} explore implicit representations-based methods to reduce decoding complexity while achieving competitive performance.

\vspace{-1em}
\paragraph{Transformer for Neural Compression}
Inspired by the effectiveness of transformer \cite{vaswani2017attention} in mainstream computer vision tasks \cite{liu2021swin,liu2022video,liu2023survey,selva2023video}, researchers have recently begun adapting Transformers for image and video compression.

Qian et al. \cite{qian2022entroformer} proposed a transformer-based entropy model for image compression, replacing the CNN-based hyperprior network \cite{minnen2018joint} with a self-attention block. 
Zhu et al. \cite{zhu2022transformer} incorporated the swin transformer \cite{liu2021swin} into a standard neural image compression framework \cite{balle2018variational, minnen2020channel} to facilitate nonlinear image transformation and entropy coding. 
Liu et al. \cite{liu2023learned} adopted a parallel transformer-CNN structure for image compression to combine the strengths of local and long-distance modeling capabilities.
To accelerate the decoding speed, Mentzer et al. \cite{mentzer2023m2t} proposed M2T image compression model to mask transformer twice.
Koyuncu et al. \cite{koyuncu2023efficient} presented a spatio-channel attention entropy model for fast image context modeling when compression.

\begin{figure*}[ht]
    \centering
    \includegraphics[width=\linewidth]{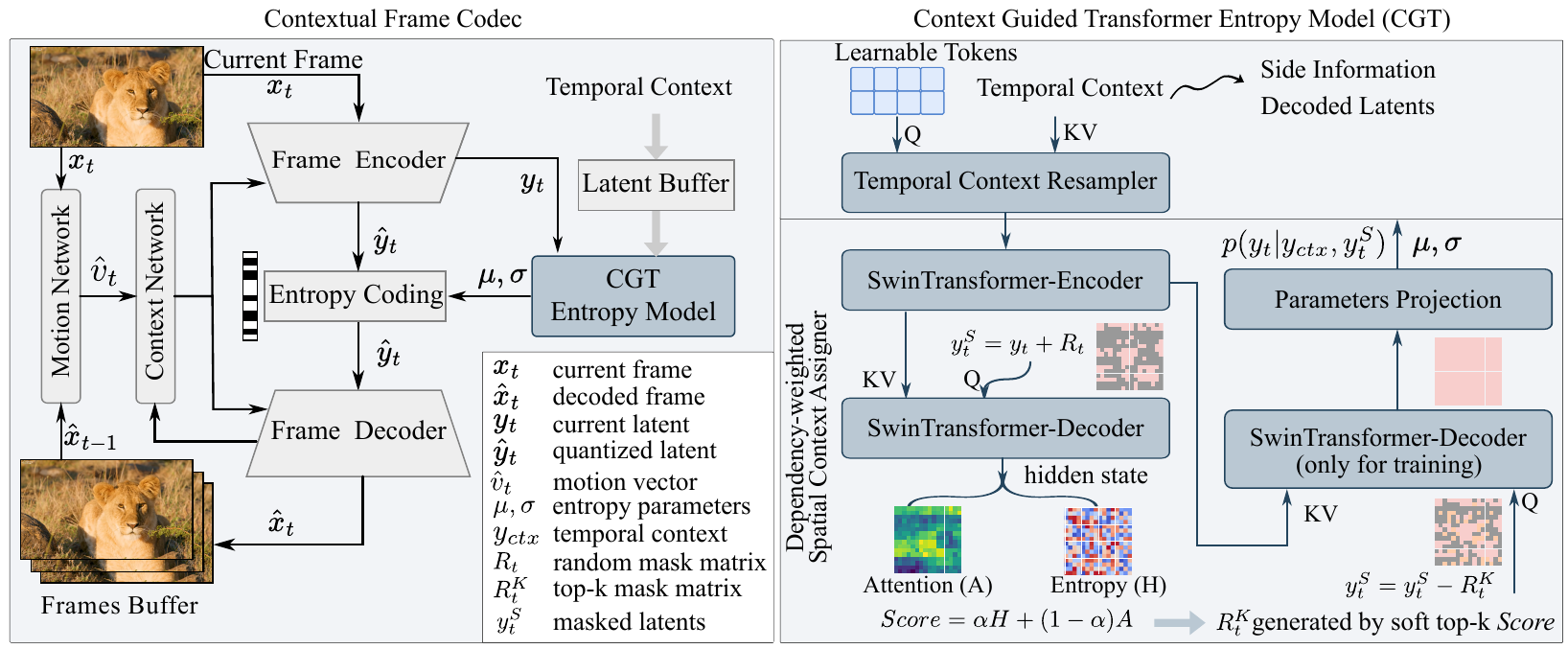}
    \vspace{-2em}
    \caption{ Overview of our video compression model. The model generates a latent representation $y_t$ of the current frame $x_t$ through a contextual frame codec.  
    Given temporal and spatial contexts, the CGT model performs probabilistic modeling to provide the PMF for entropy coding. The temporal contexts such as $y_{t-1}$ are first resampled to capture key dependencies and significantly reduce model overhead. A transformer-based teacher network generates a masked representation $y_t^M$ by modeling the importance of spatial context, guiding the student network to efficiently utilize contextual information.
    }
    \vspace{-1em}
    \label{Fig:framewowrk}
\end{figure*}

The VCT \cite{mentzer2022vct} is the first application of transformer architecture to video compression.  
Within the VCT framework, two previous frames provide temporal context, with the Vaswani transformer facilitating feature extraction. 
The concatenated feature is then combined with the current frame as spatial context, and an autoregressive mechanism is employed for conditional entropy modeling.
Chen et al. \cite{chen2023neural} introduced an optical flow-based temporal context mining module and incorporated a 3D convolution-based attention mechanism to jointly leverage spatial-temporal feature.
To overcome the unidirectional context limitations inherent in VCT's autoregressive transformer, Xiang et al. \cite{xiang2023mimt} introduced a random masked image modeling transformer to learn the spatial-temporal dependencies bidirectionally.

\vspace{-0.5em}
\section{Background and Preliminary}
Let $\{x_t\}_{t=1}^{N}$ denote a sequence of video frames and $\{y_t\}_{t=1}^{N}$ denote the corresponding latent representations, where $N$ is the video length.
Mainstream neural video compression methods follow a nonlinear transformation framework that each frame $x_t$ is mapped to $y_t$. 
Then the latents will be quantized and coded as bitstream by entropy coding.
Specifically, entropy model is designed to estimate a PMF $q(y_t)$ to close the actual PMF $p(y_t)$ of the latent representation $y_t$. 
According to information theory, the low bound of the bit-rate can be represented as a cross-entropy $H(p, q)$.

In this paper, we follow aforementioned process and focus on conditional entropy models.
The entropy model estimates the PMF conditioned on the contextual information, and its optimization objective can be written as:
\begin{align}
    \label{eq-1}
    H(p, q)&= \mathbb{E}_{\boldsymbol{y}_{t} \sim p}\left[-\log _{2} q\left(\boldsymbol{y}_{t} \mid \boldsymbol{y}_{t}^T, \boldsymbol{y}_{t}^{S},\boldsymbol{\theta}\right)\right],
\end{align}
where $\boldsymbol{y}_{t}^T$ represents the temporal contexts, $\boldsymbol{y}_{t}^{S}$ denotes the previously encoded portion of the current frame employed for spatial context, and $\boldsymbol{\theta}$ is the parameter of entropy model.

This paper aims to more effectively exploit temporal and spatial contexts for improved PMF modeling, which determines the final compression performance.

\section{Methodology}

The framework of our model is shown in Figure \ref{Fig:framewowrk}, which can be divided as a frame codec and a context guided transformer entropy model (CGT).
The frame codec leverages both inter-frame and intra-frame information to encode video frames $\{x_t\}_{t=1}^{N}$ into latent tokens $\{y_t\}_{t=1}^{N}$. 
The CGT then utilizes transmitted information, encompassing temporal context from the latent buffer and spatial context from already decoded tokens within the current frame, to compress the current latent representation $y_t$.

\subsection{Frame Codec}

In line with most existing works \cite{sheng2022temporal,xiang2023mimt}, we employ a contextual-based image encoder-decoder. This encoder-decoder consists of a CNN-based image encoder and decoder, and a temporal context mining model.
The CNN image encoder maps an RGB image of shape $(H,W,3)$ to a latent space feature map of shape $(h,w,C)$, which serves as the input to our conditional entropy model. 
On the decoding side, the quantized tokens $\{\hat{y}\}_{t=1}^{N}$ are mapped back to the image space via the image decoder to obtain the reconstructed image $\{\hat{x}\}_{t=1}^{N}$.
The temporal context mining model \cite{sheng2022temporal, guo2023enhanced} consists of a motion network and a context network. It learns temporal context from historical frames in the frame buffer while leveraging multi-scale features to enhance motion and texture representation in the spatio-temporal volume.
The process can be expressed as:
\begin{align}
    \label{eq-2}
    \nonumber
\boldsymbol{y}_{t}&=\textit{E}\big(\boldsymbol{x}_t,f(\boldsymbol{v}_t,\hat{\boldsymbol{y}}_{t-1})\big)\\
\hat{\boldsymbol{x}}_{t}&=\textit{D}\big(\hat{\boldsymbol{y}}_t,f(\boldsymbol{v}_t,\hat{\boldsymbol{y}}_{t-1})\big),
\end{align}
where $v_t$ is the estimated optical flow, and $f$ represents the temporal context mining module, $E$ and $D$ denote the encoder and decoder, respectively.

\begin{figure}[t]
    \centering
    \includegraphics[width=\linewidth]{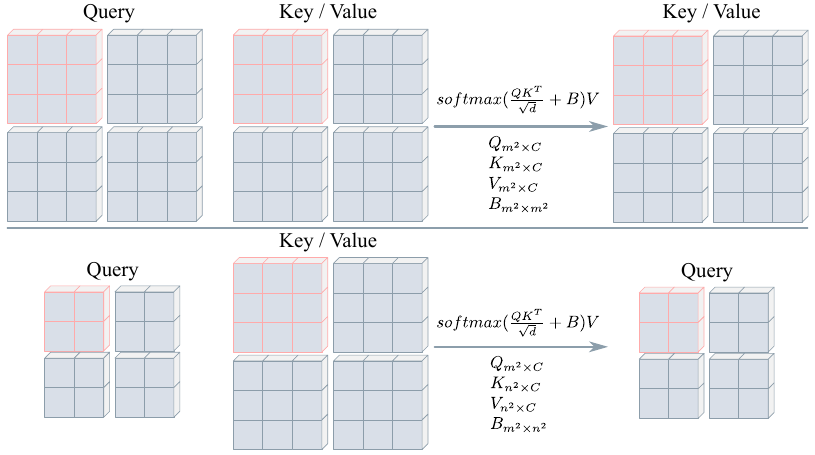}
    \caption{ Comparison of window self-attention (top) and window cross-attention (bottom), where TCR utilizes window cross-attention to compress temporal latents through learnable queries.
    }
    \label{Fig:window}
\end{figure}

\subsection{Context Guided Transformer Entropy Model}
\paragraph{Temporal Context Resampler}

Temporal context facilitates entropy models in capturing the inter-frame dependencies essential for encoding temporal redundancies in videos. Intuitively, not all contextual information holds equal importance, and as information increases, decoding speed is significantly affected. 

The CGT model includes a swin transformer-based temporal context resampler that extracts effective features from diverse types (e.g., historical frames, hyper priors) and scales of temporal context. As shown in the right of Figure~\ref{Fig:framewowrk}, the temporal context resampler employs a set of trainable window queries as input and produces fixed-length tokens via swin transformer blocks. First, we pre-define a small learnable query and fuse it with the larger temporal context via window cross-attention. Through the powerful modeling capability of swin transformer, key features are extracted from the key-value pairs. These contextual tokens are concatenated and fed into a transformer for probabilistic distribution modeling of conditioning information. 

Figure~\ref{Fig:window} compares the window self-attention and window cross-attention, where the latter allows attention computation across windows of varying sizes. In our TCR module, the small and learnable queries are divided into the same number of windows as the temporal context, enabling localized compression of effective information within each window for temporal latent representation.

\paragraph{Dependency-weighted Spatial Context Assigner}
The CGT also includes a dependency-weighted spatial context assigner (DWSCA) to explicitly model position dependency of spatial context.
As illustrated in the right of Figure~\ref{Fig:framewowrk}, the dependency-weighted spatial context assigner comprises a swin transformer encoder and a teacher-student network.
The compact temporal features are obtained via the temporal context resampler, then fused by the swin transformer encoder, whose output serves as the temporal context for both the teacher and student networks. 
The teacher-student network consists of a pair of swin transformer decoders with shared parameters. 
During training, a teacher-student network is used for explicit spatial dependency modeling. Since share parameters, only one is required during inference to generate dependency relationships and perform PMFs estimation, ensuring an efficient decoding process.

Specifically, during training, we employ a teacher-student network to simulate the step-by-step decoding process. The teacher network takes the partially decoded content of the current frame as input and applies the window attention mechanism of the swin transformer to generate an attention map, representing the importance of each spatial position, and an entropy map, indicating the certainty of each spatial position. To balance spatial importance and certainty, we compute a weighted combination of these two maps, producing a dependency score that quantifies the strength of spatial dependencies, as shown in Equation~\ref{eq-weight}:
\begin{align}
    \label{eq-weight}
    Score = \alpha H+ (1-\alpha)A,
\end{align}
where $A$ denotes the normalized attention map and $H$ denotes the normalized entropy map.
By performing top-k ranking, we identify the most contextually dependent regions, which are then decoded and serve as context for the progressive decoding process in the student network. The student network receives both the partially decoded content of the current frame and the decoded content from the teacher network, enabling it to iteratively refine and reconstruct the full frame while maintaining consistency in spatial dependency modeling.
Notably, we employ soft top-k \cite{xie2020differentiable, petersen2022differentiable} as a differentiable approximation for the ranking process, enabling gradient propagation during training.

Although the above process simulates the step-by-step decoding procedure, the decoded content of the current frame in the teacher network is not explicitly predefined. To address this, we draw inspiration from image masked modeling \cite{he2022masked,xiang2023mimt} and apply random masking to the input latents, where the unmasked regions represent the already decoded content of the current frame.
The masked encoding and decoding process can be described as follows:
\begin{align}
    \label{eq-2}
R_{\text {mask }}\left(\boldsymbol{y}_{t}\right)=\mathbb{E}_{\boldsymbol{y}_{\boldsymbol{t}} \sim p}\left[-\log _{2} q\left(\boldsymbol{y}_{t} \mid \boldsymbol{y}_{t}^T, \boldsymbol{y}_{t}^{{\mathbf{S}}}\right)\right],
\end{align}
where $y_t^T$ denotes the resampled temporal context, $y_t^S$ represents the already decoded portion of the current latent, serving as the spatial context for the tokens undecoded. 
During training, $y_t^S$ can be simulated by applying a random masking matrix to the latent representations, i.e., $y_t+M$, where $M$ is the random masking matrix.
After the masking process, teacher network generates attention map and entropy map based on the latent with random masking matrix. The decoding process of the teacher network can be represented by removing the mask from designated positions. Then we can estimate the PMF of the current frame using student network. The output of the student network is passed through a parameter projection layer to obtain the parameter estimates of the PMF as shown in Figure~\ref{Fig:framewowrk}.

\subsection{Decoding Process}
The CGT entropy model is designed for step-by-step decoding. we first need to define the decoding step size and the number of tokens decoded per step. Following \cite{xiang2023mimt}, we adopt an 8-step scheduling strategy based on a sinusoidal function.
The CGT decoding process is elucidated in Algorithm~\ref{alg:decoding}. We commence by decoding prior information such as optical flow and hyperpriors from the bitstream. Subsequently, we progressively decode the current frame, guided by temporal context and attention mechanisms. Each decoding step takes into account the dependency-weighted score of token positions, prioritizing the decoding of tokens with high importance scores and high certainty scores.

\begin{algorithm}[b]
    \caption{Decoding process of CGT model}
    \textbf{Input:} Context $y_{t_p},y_{h_p},\hat{y}_{t-1}$, bit-stream, decoding step $n$.\\
    \textbf{Output:} $\hat{x}_t$.
    \begin{algorithmic}[1]
    \State Decoded $\hat{v}_t$ and $\hat{z}_t$ from bit-stream;
    \State Decoded the temporal-prior $y_{t_p}$ and the hyperprior $y_{h_p}$;
    \While{$i < n$}
    
    \State  Resample $y_{t_p}, y_{h_p}$, and $\hat{y}_{t-1}$ using TCR;
    \State  Fuse the resampled temporal context using the swin transformer encoder;
    \State  Perform cross-attention between the fused context and $y_t^S$ using the swin transformer decoder;
    \State  Generate attention map and entropy map based on cross-attention output;
    \State  Perform dependency-weighted processing and apply top-k selection;
    \State  Decode distribution of the top-k position;
    \EndWhile
    \State Decode the current frame using entire distribution parameters.
    \end{algorithmic}
    \label{alg:decoding}
\end{algorithm}

\subsection{Model Parameter and Loss Function}
To maintain consistency in the overall structure of the model, we choose to use the same swin transformer block with temporal context resampler as the decoder. The encoder is obtained by replacing the cross-window attention with self-attention. The dimension of the transformer is set to 768, the dimension of the image encoder-decoder is set to 96, the teacher and student networks are each set to 4 transformer blocks, and the context sampler and transformer encoder are each set to 2 transformer blocks.

The loss function of the CGT model is
\begin{align}
    \label{eq-object}
\mathcal{L}_{\mathrm{RD}}=\underbrace{R\left(\hat{\boldsymbol{y}}_{t}\right)+R\left(\hat{\boldsymbol{z}}_{t}\right)+R\left(\hat{\boldsymbol{v}}_{t}\right)}_{\text {bit-rate }}+\lambda \cdot \underbrace{d\left(\boldsymbol{x}_{t}-\hat{\boldsymbol{x}}_{t}\right)}_{\text {distortion }},
\end{align}
where $R$ is the bit-rate term, $d$ is distortion term, and the $\mathcal{L}_{CE}$ denotes the masked image model loss.
We set coefficient $\lambda$ to ${256,512,1024,2048}$ for RD trade-off.

This paper employs the latents of the previous frame $y_{t-1}$, the hyper-prior $y_{h_p}$, and the temporal-prior $y_{t_p}$ \cite{li2022hybrid} as the temporal contexts. The optimization objective can be reformulated as follows:
\begin{align}
    \label{eq-object}
    R\left(\boldsymbol{y}_{t}\right)=\mathbb{E}_{\boldsymbol{y}_{\boldsymbol{t}} \sim p}\left[-\log _{2} q\left(\boldsymbol{y}_{t} \mid \boldsymbol{y}_{t-1}, \boldsymbol{y}_{h_p}, \boldsymbol{y}_{t_p}, \boldsymbol{y}_{t}^{{\mathbf{S}}}\right)\right].
\end{align}

\section{Experiments}

\subsection{Experimental Setup}
\textbf{Datasets.} Consistent with most works, Vimeo-90k \cite{xue2019video} is used as the training set in this paper. The dataset consists of 91,701 video sequences, each containing 7 frames with a resolution of $448 \times 256$. We randomly crop them to $265 \times 256$ and augment the data by random flipping. The test set includes MCL-JCV \cite{wang2016mcl}, UVG \cite{mercat2020uvg}, and HEVC class B, all with a resolution of  $1920 \times 1080$. Following \cite{xiang2023mimt}, we center-crop the test set to  $1792 \times 1024$ to be divisible by 64.

 \textbf{Metrics.}We employ PSNR and MS-SSIM in the RGB color space as an evaluation metric for distortion. The rate is measured by bit per pixel (bpp).

\textbf{Baselines.} 
 the performance of video compression is influenced by both the frame encoder and the entropy model. Notably, this study focuses on conditional entropy model modeling, our primary comparison is with different conditional entropy models. We adopt the frame codec from the DCVC model as our backbone.
Our proposed CGT model is compared with conventional codecs and neural video compression methods.
Conventional codecs include H.265 and H.266, corresponding to HM and VTM, respectively. 
For neural video compress methods, we directly report the results of the following baselines:  MIMT \cite{xiang2023mimt}, VCT \cite{mentzer2022vct}, DMC \cite{li2022hybrid}, DCVC \cite{li2021deep}, DCVC-DC \cite{li2023neural}, FVC \cite{hu2021fvc}, and C2F \cite{hu2022coarse}.

\begin{table*}[ht]
\centering
\begin{tabular}{ccrlrlrl} 
\toprule
\multirow{2}{*}{Model}   & \multirow{2}{*}{BD-Rate (\%)} & \multicolumn{6}{c}{Time consumption}                                                                                                                             \\ 
\cmidrule{3-8}
                         &                               & \multicolumn{2}{l}{Entropy modeling time (ms)~}     & \multicolumn{2}{l}{Encoding time (ms)}              & \multicolumn{2}{l}{Decoding time (ms)}               \\ 
\midrule
MIMT(random)-w/o \cite{xiang2023mimt}        & \textbf{0}                             & 1,226 & \multirow{2}{*}{\textbf{$\downarrow$ 65\%}} & 1,591 & \multirow{2}{*}{\textbf{$\downarrow$ 34\%}} & 1,480 & \multirow{2}{*}{\textbf{$\downarrow$ 36\%}}  \\
\textbf{MIMT(random)-w } & +2.2                 & 427   &                                             & 1,058 &                                             & 943   &                                              \\ 
\midrule
CGT-w/o                  & 0                             & 1,305 & \multirow{2}{*}{\textbf{$\downarrow$ 63\%}} & 1,682 & \multirow{2}{*}{\textbf{$\downarrow$ 35\%}} & 1,576 & \multirow{2}{*}{\textbf{$\downarrow$ 38\%}}  \\
\textbf{CGT-w}           & {+1.8}                 & 488   &                                             & 1,073 &                                             & 984   &                                              \\
\bottomrule
\end{tabular}
    \caption{\small Ablations on the temporal context resampler module, where the anchor is the plain swin-transformer block. The CGT-w denotes the CGT model with temporal context resampler. The decoding step for all models is set to 8. All experiments were conducted with an NVIDIA A800 GPU.}
    \label{tab:freq}
\end{table*}

\begin{table*}[tbp]
    \centering
    \begin{tabular}{ccccc}
    \toprule
            Model &  MCL-JCV dataset (\%) & UVG dataset (\%) & HEVC-B dataset (\%) &Average (\%)\\
         \midrule
         \textbf{CGT-DWSCA (Our)}&\textbf{ -11.3}& \textbf{-7.8} & \textbf{-14.6}& \textbf{-11.2}\\
         CGT-min-entropy \cite{xiang2023mimt} & 0 &0 & 0& 0\\
         CGT-checkerboard \cite{he2021checkerboard} & 17.7 &15.1 & 19.2& 17.3\\
         CGT-block-autoregressive \cite{mentzer2022vct}& 19.3 &16.6 & 22.8& 19.5\\

        \bottomrule
    \end{tabular}
    \caption{\small Ablations on the different entropy model, where the anchor is set as the MIMT model, which using minimal entropy principle for decoding\cite{xiang2023mimt}. The first row denote the CGT model with our dependency-weighted spatial context assigner methods.}
    \label{tab:mask}
\end{table*}

\subsection{Evaluation of Temporal Context Resampler}

To demonstrate the effectiveness of the temporal context resampler, we conduct ablation experiments on our CGT entropy model and the MIMT \cite{xiang2023mimt} with random mask, respectively. The baseline model is set to extract temporal context features through a plain swin transformer block with window self-attention.

The results in Table~\ref{tab:freq} indicate that while incorporating temporal context into the CGT model increases the BD rate by 1.8\%, it remarkably reduces entropy modeling time (i.e., the time required to estimate the PMF) by 63\%. Furthermore, under the 8-step progressive decoding baseline, introducing the TCR module reduces encoding time and decoding time by 35\% and 38\%, respectively.
Similar results are observed in the MIMT \cite{xiang2023mimt} model. Although incorporating temporal context into the random masking model improves the BD rate by 2.2\%, it significantly reduces the entropy modeling time 65\%.
Furthermore, temporal context resampler achieves a reduction of over 34\% in both encoding and decoding time compared to the plain swin-transformer.

These results demonstrate that the proposed temporal context tesampling module outperforms standard swin transformer blocks. The TCR module effectively enhances the quality of temporal context information while reducing subsequent processing time by leveraging compact, learnable queries for resampling, all without significantly compromising the model’s encoding capability. Moreover, this approach exhibits strong generalization across different swin transformer-based models.

\subsection{Evaluation of Dependency-Weighted Spatial Context Assigner}
\paragraph{Effectiveness Analysis}

Our dependency-weighted spatial context assigner fundamentally emphasizes the dependancy of spatial context by employing a random number of masks with relatively fixed positions.

The teacher-student network simulates the progressive decoding of video frames, where fixed mask positions guide the identification of informative contextual regions.
The teacher generates weighted attention maps and entropy maps to measure dependencies, providing the most relevant context for the student. We emphasize explicit modeling of spatial context dependencies to distinguish our approach from previous methods. To demonstrate its effectiveness, we compare it with autoregressive decoding \cite{mentzer2022vct}, checkerboard decoding \cite{he2021checkerboard}, and minimal entropy decoding \cite{xiang2023mimt}.

Table~\ref{tab:mask} presents the BD-rate results of various spatial context modeling approaches, using the minimal entropy decoding method as the anchor.
As shown in the table, the BD-rate performance of both autoregression and checkerboard is lower than that of the  minimal entropy decoding strategy, which has been verified in the literature \cite{xiang2023mimt}. In contrast, the first and second rows indicate that our dependency-weighted spatial context assigner strategy outperforms minimal entropy decoding in terms of rate-distortion performance. This demonstrates that modeling the importance of spatial context positions has a positive impact on improving the efficiency of context utilization.

\paragraph{Analysis of the Impact of Weighting Coefficient}

\begin{table}[t]
\centering
\setlength{\tabcolsep}{11pt}
\begin{tabular}{l|l|lll} 
\toprule
\multicolumn{1}{c|}{} & \multicolumn{1}{c|}{} & \multicolumn{3}{c}{$\alpha$} \\
\cmidrule(lr){3-5}
\multicolumn{1}{l|}{$\lambda$} & Metric & 1     & 0     & 0.5    \\ 
\hline
\multirow{2}{*}{256} & PSNR   & 35.3  & 35.88 & 35.82  \\
                     & Bpp    & 0.017 & 0.019 & 0.018  \\ 
\hline
\multirow{2}{*}{512} & PSNR   & 36.74 & 37.17 & 37.01  \\
                     & Bpp    & 0.035 & 0.037 & 0.036  \\
\bottomrule
\end{tabular}
\caption{Impact of different $\alpha$ values on model performance.}
\label{tab:lambda}
\end{table}

\begin{table}[t]
\centering
\setlength{\tabcolsep}{8pt}
\begin{tabular}{l|l|lll} 
\toprule
\multicolumn{1}{c|}{} & \multicolumn{1}{c|}{} & \multicolumn{3}{c}{$\alpha$} \\
\cmidrule(lr){3-5}
\multicolumn{1}{l|}{Setting} & Metric & 1     & 0     & 0.5    \\ 
\hline
\multirow{2}{*}{Explict} & PSNR   & \textbf{35.3}  & \textbf{35.88} &\textbf{ 35.82}  \\
                     & Bpp    & \textbf{0.017} &\textbf{ 0.019} & \textbf{0.018}  \\ 
\hline
\multirow{2}{*}{Proxy task} & PSNR    & 35.21 & 35.82 & 35.58  \\
                     & Bpp   & \textbf{0.017} & 0.021 & 0.019  \\
\bottomrule
\end{tabular}
\caption{Result of the impact of spatial context explicit modeling.}
\label{tab:explicit}
\end{table}

We analyze the impact of spatial position importance and certainty by adjusting the weighting coefficient. For simplicity, we set $\alpha=0, 1$, and 0.5, corresponding to three scenarios: the model focusing solely on spatial importance, focusing solely on spatial certainty, and equally considering both factors, respectively.
We evaluate the results of the aforementioned settings on the MCL-JCV dataset, where the result is shown in Table~\ref{tab:lambda}.

When $\alpha=0$, the model primarily focuses on the Attention Map, emphasizing the importance of spatial positions. Consequently, the reconstruction loss decreases, as the model prioritizes decoding more informative regions.
In contrast, when $\alpha=1$, the model primarily relies on the Entropy Map, favoring spatial certainty. As a result, the bitrate consumption is reduced, as the model decodes more predictable regions first, optimizing entropy coding efficiency.

We simply set the coefficient to 0.5, allowing the model to jointly consider both the importance and certainty of spatial positions, thereby balancing their influence on the decoding process.
Note that the model applies weighting to normalized attention and normalized entropy, meaning that 
$\alpha=0.5$ represents an equal emphasis on both factors.

\paragraph{Analysis of the Impact of Explicit Modeling}
To intuitively demonstrate the benefits of explicit spatial dependency modeling, we compare the performance of models trained using a proxy task versus our proposed explicit modeling approach, as shown in Table~\ref{tab:explicit}.
The results indicate that while proxy-task-based modeling introduces training diversity, it suffers from training-inference mismatch. As the model passively adapts to random masking during training but actively selects optimal paths during inference, it struggles to fully cover all possible minimum-entropy paths, leading to suboptimal performance.

In contrast, our teacher-student network combined with soft top-k selection explicitly models spatial context dependencies, ensuring greater alignment between training and inference. This consistency enables more effective context selection, ultimately leading to improved compression efficiency and reconstruction quality.

\subsection{Generalization Capability}

\begin{figure}[t]
    \centering
    \includegraphics[width=0.95\linewidth]{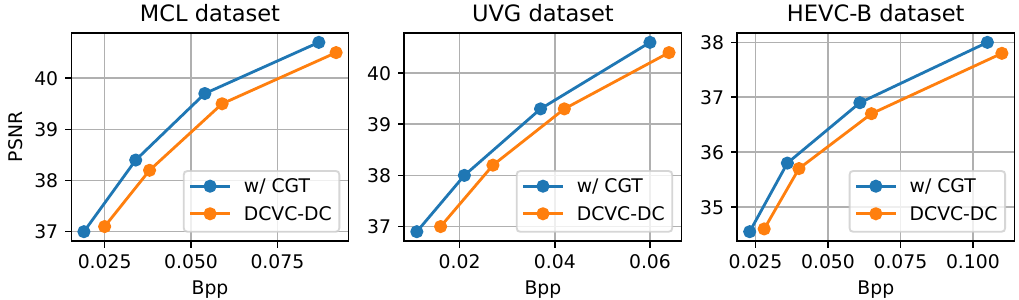}
    \caption{\small Generalization capability of our CGT entropy model based on DCVC-DC framework.}
    \label{Fig:plt-gen}
    \vspace{-1em}
  \end{figure}

To demonstrate the generalization capability of the proposed CGT model, we replace the frame codec from DMC to DCVC-DC \cite{li2023neural} and compare it against state-of-the-art models. The experimental results are shown in Figure~\ref{Fig:plt-gen}.

The results indicate that even when using a more advanced frame codec, our CGT model remains highly adaptable and consistently outperforms the baseline in entropy coding efficiency, demonstrating its superior capability in conditional entropy modeling.
\subsection{Rate-Distortion Performance Comparisons}

\begin{figure*}[htb]
    \centering
    \includegraphics[width=0.95\linewidth]{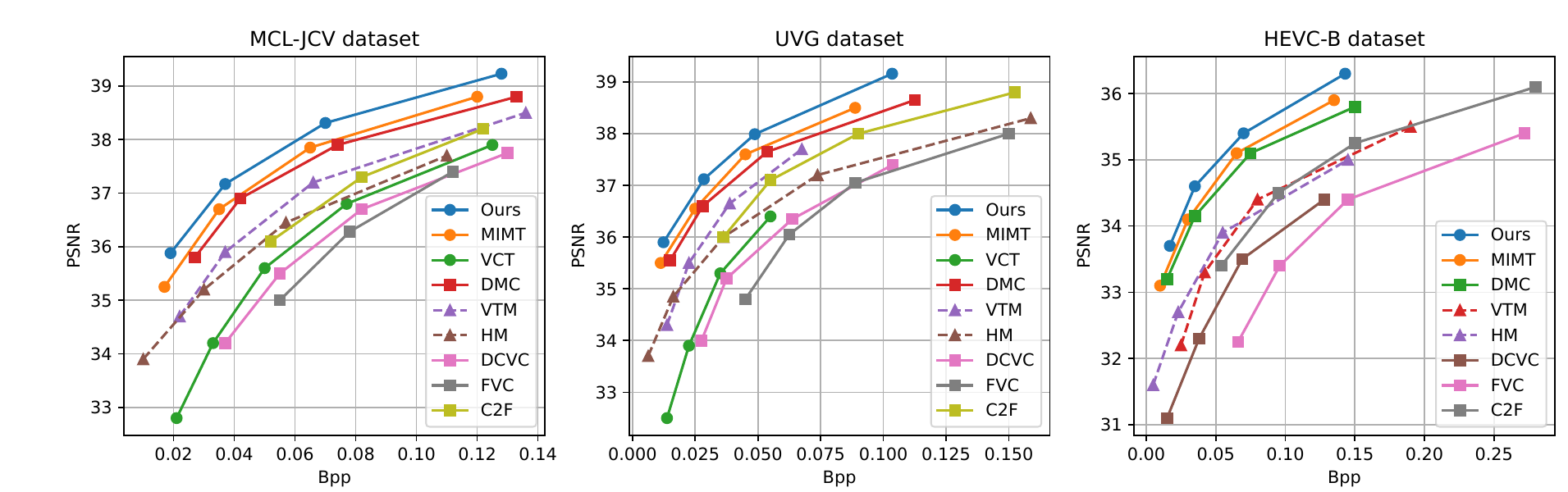}
    \includegraphics[width=0.95\linewidth]{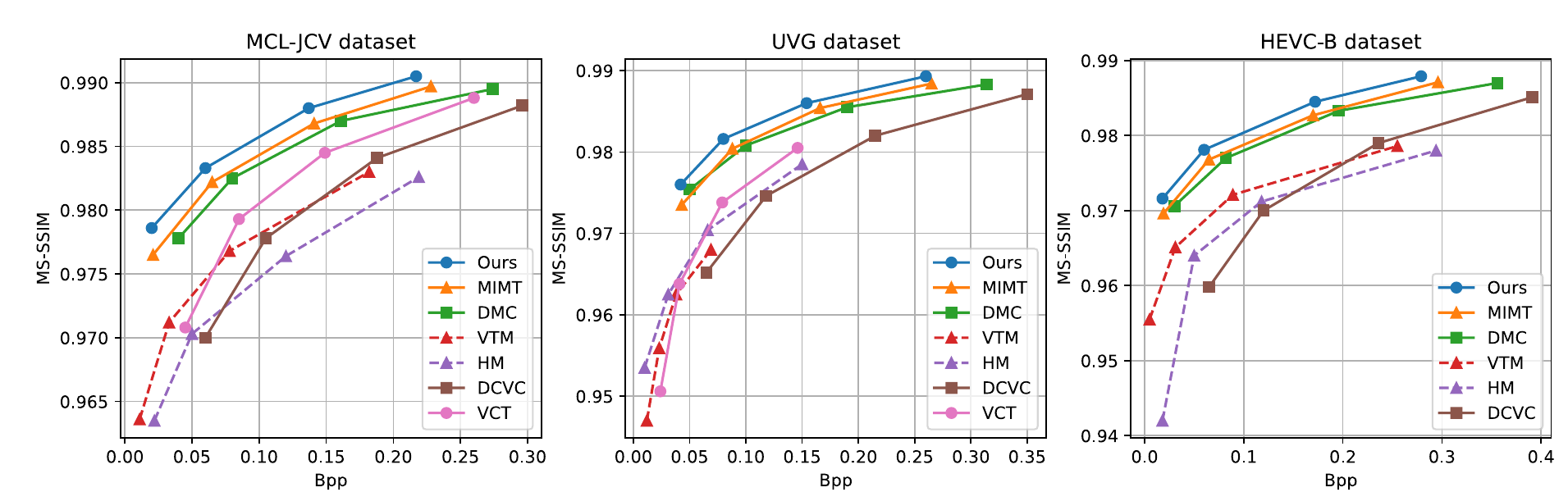}
    \vspace{-1em}
    \caption{\small RD Performance on UVG, MCL-JCV, and HEVC-B Datasets. Distortion is evaluated using PSNR in the first row of images, whereas MS-SSIM is employed as the distortion metric in the second row.}
    \label{Fig:plt}
    \vspace{-1em}
  \end{figure*}

The Figure~\ref{Fig:plt} illustrates the rate-distortion (RD) performance curves of our proposed method compared to baseline methods on three test datasets.
Our model consistently achieves lower bitrate costs at the same distortion levels, outperforming conditional entropy-based methods such as MIMT and VCT.
This highlights the CGT model's exceptional ability to effectively utilize spatiotemporal context and obtain more accurate PMF estimates.

In addition, to quantitatively compare the performance of the models, we use VTM as the anchor and calculate the BD-rate between the remaining models and VTM.
BD-rate \cite{bjontegaard2001calculation} is a commonly used metric for comparing the performance of video codecs. It is typically used to evaluate the compression efficiency of different codecs. It measures the change in bitrate achieved by using one codec compared to another at the same video quality.
A lower BD-rate indicates that the model has better rate-distortion performance compared to the anchor model.
Table~\ref{tab:bdrate} and \ref{tab:bdrate-2} show the quantitative comparison results of our CGT model with the baseline models. Some of the original data comes from \cite{xiang2023mimt}.
The random masking modeling scheme shows an average BD-rate decrease of 22.7\% on the three datasets, while our model dropped by 47.7\%.

The results demonstrate that our CGT model achieves the most significant BD-rate improvement over the VTM anchor. In addition, compared to prior methods based on random masking strategies, our model yields superior performance, indicating a more effective utilization of both temporal and spatial context for entropy modeling.

\vspace{-0.5em}
\begin{table}[tbp]
    \centering
    \begin{tabular}{lcccc}
    \toprule
               & \textbf{MCL-JCV} & \textbf{UVG} & \textbf{HEVC-B} & \textbf{Average} \\
    \midrule
    \textbf{FVC}    & 71    & 108.6  & 108.3  & 95.9  \\
    \textbf{DCVC}   & 66.1  & 95.8   & 49.8   & 70.5  \\
    \textbf{C2F}    & 19.6  & 16     & 14.2   & 16.6  \\
    \textbf{VTM}    & 0     & 0      & 0      & 0     \\
    \textbf{DMC}    & -24.5  & -26.1  & -49.4   & -33.3  \\
    \textbf{MIMT}   & -33.0  & -34.9  & -57.1  & -41.7 \\
    \textbf{CGT}    & -43.8 & -45.5  & -62.5  & -50.6 \\
    \bottomrule
    \end{tabular}
    \vspace{-0.5em}
    \caption{\small BD-rate (\%) calculated by PSNR, where the VTM is used for anchor. The lower the BD-rate, the better the compression performance. Some of the original data comes from \cite{xiang2023mimt}.}
    \label{tab:bdrate}
    \vspace{-1em}
\end{table}

\begin{table}[tbp]
    \centering
    \begin{tabular}{lccccccc}
    \toprule
    & \textbf{MCL-JCV} & \textbf{UVG} & \textbf{HEVC-B} & \textbf{Average} \\
    \midrule
    FVC   & 15.0 & 10.1 & 60.9 & 28.6  \\
    DCVC  & 4.7  & 33.6 & 31.0 & 23.1  \\
    VTM   & 0    & 0    & 0    & 0     \\
    C2F   & -9.1 & -20.1 & 12.7 & -5.5  \\
    DMC   & -54.3 & -55.1 & -56.6 & -55.4 \\
    MIMT  & -63.5 & -67.7 & -64.8 & -65.3 \\
    \textbf{CGT} & -74.6 & -73.8 &-75.7 & -74.7 \\
    \bottomrule
    \end{tabular}
    \vspace{-0.5em}
    \caption{\small BD-rate (\%) calculated by SSIM, where the VTM is used for anchor. The lower the BD-rate, the better the compression performance. Some of the original data comes from \cite{xiang2023mimt}.}
    \label{tab:bdrate-2}
    \vspace{-1em}
\end{table}
\section{Conclusion}
This paper introduces the Context-Guided Transformer (CGT) entropy model, which improves PMF estimation in video compression by effectively leveraging spatio-temporal context.
CGT employs a temporal context resampler that uses latent queries and transformer encoders to extract critical temporal features while reducing computational overhead.
Meanwhile, a teacher-student framework explicitly models spatial context dependencies: the teacher generates attention and entropy maps from randomly masked inputs to guide the student in selecting the most informative spatial positions.
During inference, only the student is used to predict undecoded tokens based on high-dependency context, achieving accurate prediction with low complexity.
Experimental results show that CGT reduces entropy modeling time by 65\% and achieves an 11\% BD-rate improvement over prior state-of-the-art models, demonstrating its ability to balance compression efficiency with computational complexity.

\newpage
\section*{Acknowledgements}
We thank EIT and IDT High Performance Computing Center for providing computational resources for this project. This work was supported by the  2035 Key Research and Development Program of Ningbo City under Grant No.2024Z127.

{
    \small
    \bibliographystyle{ieeenat_fullname}
    \bibliography{ref}
}
\newpage
\clearpage
\appendix
\setcounter{figure}{0}   
\setcounter{table}{0}
\setcounter{equation}{0}
\setcounter{algorithm}{0}

\section{Network Structure}
\paragraph{Image encoder, decoder}
In this paper, we focus on conditional entropy modeling based on a well-designed frame codec.
The frame codec employed in this work follows a contextual encoding backbone \cite{sheng2022temporal,guo2023enhanced}, comprising a frame encoder, decoder, and a temporal context mining module.
The frame encoder employs 4 strided convolutional layers achieving a 16× downsampling factor, while the frame decoder utilizes 4 strided transposed convolutions for upsampling to reconstruct the input frame.
The temporal context mining module learns multi-scale temporal contexts directly from the propagated feature generated by the previously decoded frame $\hat{x}_{t-1}$ and the optical flow $\hat{v}_t$.
Multi-scale contexts are used as a reference for both the encoder and decoder.
The frame codec map the input frame to a sequence of tokens with dimension of 96 in latent space.
Consistent with previous works \cite{li2022hybrid,xiang2023mimt}, we employ Anchor model \cite{cheng2020learned} as the encoder for I-frames and set the GoP to 32.
We train the frame encoder $E$ and decoder $D$ with a hyper-prior approach \cite{balle2018variational}.
The well-trained model will serve as the basis for the context-guided transformer entropy model.

To demonstrate the generalization capability of our conditional entropy model, we additionally employ DCVC-DC, proposed by Li et al. \cite{li2023neural}, as the frame encoder backbone. The primary distinction of DCVC-DC with the above frame codec lies in the improvements to the context network, which differentiate it from our main experimental setup. For further details, please refer to Li et al. \cite{li2023neural}.

\paragraph{Transformer entropy model}

\begin{figure*}[t]
    \centering
    \includegraphics[width=\linewidth]{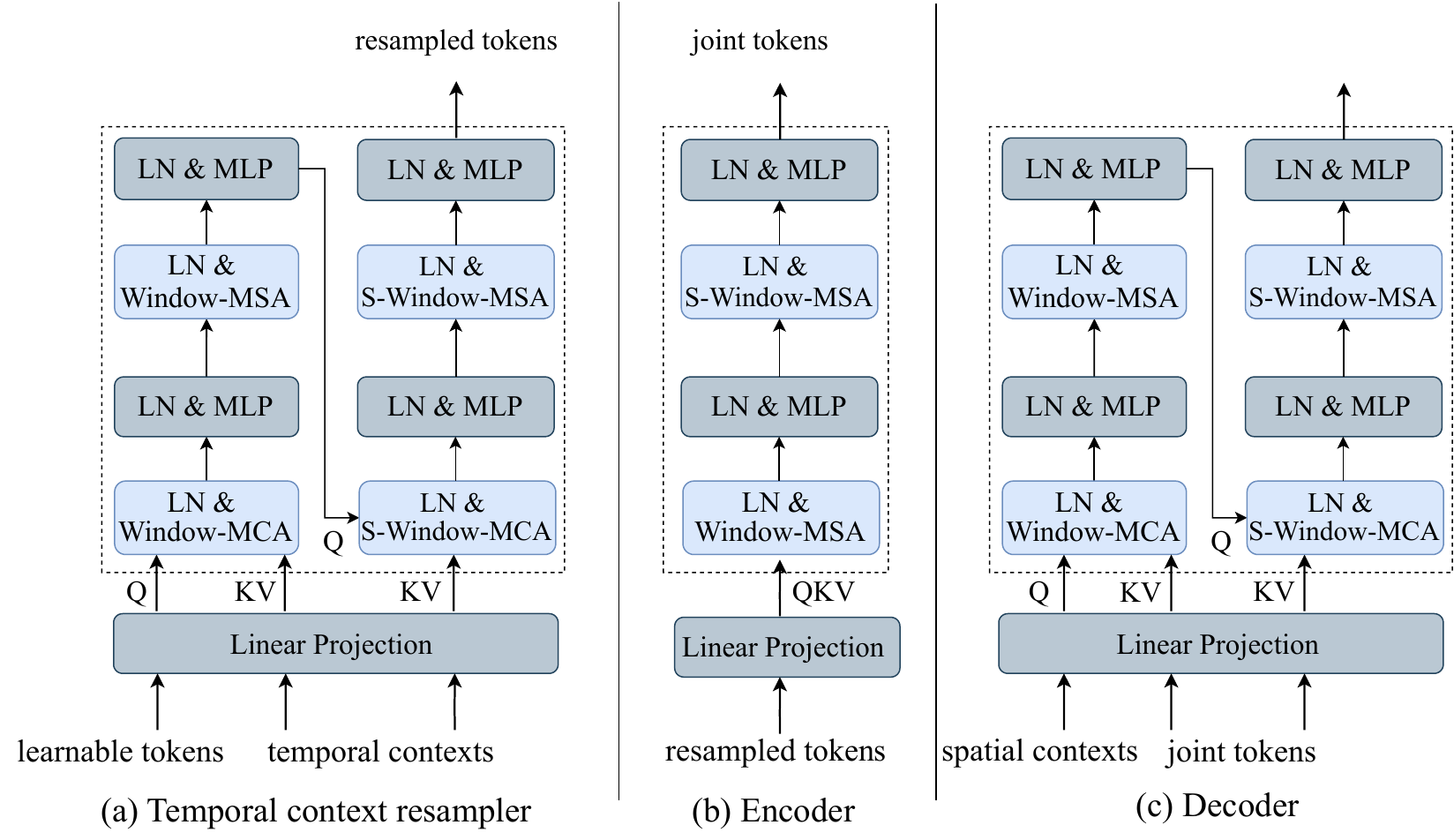}
    \caption{\small Detailed structure of the components in our CGT model. The TCR and decoder employ the same architecture for concise. In practice, the number of the temporal context resampler is set as 1, the swin transformer encoder have 2 blocks, and the swin transformer decoder have 4 blocks. }
    \label{Fig:transformer}
  \end{figure*}

As described in the main text, the CGT model consists of temporal context resampler (TCR) and an attention-guided masked model.
The attention-guided mask model consists of an encoder and a decoder based on a teacher-student network architecture.
To maintain a concise and reusable structure, both the TCR and decoder employ the same architecture. The detailed structure is illustrated in Fig \ref{Fig:transformer}, where the core of these modules is the swin-transformer block.
The swin-transformer block alternates between using window multi-head self-attention (Window-MSA) and shift-window multi-head self-attention (S-Window-MSA) layers.
For TCR and the decoder, we introduce the window multi-head cross-attention (Window-MCA) and shift window multi-head cross-attention (S-Window-MCA) into the original Swin Transformer block to handle queries and key-value pairs from different sources. The cross-attention and self-attention mechanisms are applied alternately.

The input frame with $(256,256,3)$ is mapped to latent space with $(16,16,96)$ at first using the frame codec.
Then the latents $y_{t-1},$ hyper-prior $y_{h_p},$ and temporal-prior $y_{t_p}$ are utilized as temporal contexts.
The temporal context such as $y_{t-1}$ is resampled through TCR to obtain a corresponding compact representation. The scale of resampling is controlled by the shape of learnable quires, which is set to (1, 256, 768) in this paper.
Then these temporal contexts are concatenated with the swin-transformer encoder to generate joint tokens, which are served as the key and value inputs of swin-transformer decoder.

\section{Training and Inference}
\paragraph{Training details}
The loss function of the CGT model is
\begin{align}
    \label{eq-object}
\mathcal{L}_{\mathrm{RD}}=\underbrace{R\left(\hat{\boldsymbol{y}}_{t}\right)+R\left(\hat{\boldsymbol{z}}_{t}\right)+R\left(\hat{\boldsymbol{v}}_{t}\right)}_{\text {bit-rate }}+\lambda \cdot \underbrace{d\left(\boldsymbol{x}_{t}-\hat{\boldsymbol{x}}_{t}\right)}_{\text {distortion }},
\end{align}
where $R$ is the bit-rate term, $d$ is distortion term.
We set coefficient $\lambda$ to ${256,512,1024,2048}$ for RD trade-off.
We use straight-through estimator (STE) to enable gradient propagation through quantization operations during training.

Due to observed instability when training from scratch, this work adopts a three-stage training approach in this work. First, we train the frame encoder-decoder using adjacent frames consisting of I-frames and P-frames. This process follows the standard training procedure for image codec, utilizing hyperprior training with the RD loss function. This stage focuses primarily on the reconstruction performance of the model. We initialize part of parameters with the weights of DMC \cite{li2022hybrid} and then fine-tune for 500K steps. This stage emphasizes the distortion performance of the model.
In the second stage, we use three consecutive frames as model input and freeze the frame codec parameters, training only the CGT conditional entropy model for 1M steps. This stage continues to train the model using adjacent frames consisting of one I-frame and two P-frames.
Finally, in the third stage, we incorporate the full video data and perform end-to-end fine-tuning to reduce the cumulative error between P-frames.

\paragraph{Analysis training paradigm}
In the main text, we analyzed the decoding processes of existing conditional entropy models and identified a common limitation: the lack of explicit modeling of spatial context dependencies. In this section, we further examine the training paradigms of different conditional entropy models, highlighting why our CGT model effectively captures and explicitly models the dependency within spatial context.

For autoregressive-based \cite{mentzer2022vct} and checkerboard-based \cite{he2021checkerboard} methods, the training and inference processes are strictly aligned, ensuring a consistent decoding strategy. However, this rigid structure prevents the model from differentiating the importance of contextual information at either stage, as it treats all available contexts equally without explicitly prioritizing more informative regions. 
MIMT \cite{xiang2023mimt} adopts a minimum-entropy principle for decoding and trains the model using a random masking proxy task. Since the model passively adapts to random masking during training while actively selecting the optimal path during inference, this approach, despite increasing training diversity, struggles to fully cover all possible minimum-entropy paths at inference. This limitation is particularly pronounced in complex videos with dynamically changing local features, where the optimal entropy-minimizing trajectory varies significantly.

We draw inspiration from the aforementioned training paradigms and introduce a teacher-student network along with a soft top-k strategy within the proxy task. During training, the teacher model selects the most contextually relevant regions for prioritized decoding, which then serve as the context for the student network's decoding process. This approach not only explicitly models the importance of spatial context but also ensures consistency between training and inference, bridging the gap between the two stages.

\paragraph{Training of CGT model}
The training process of our CGT conditional entropy model is illustrated in Algorithm~\ref{alg:train}.

\begin{algorithm}
    \caption{Training process of CGT model}
    \textbf{Input:} $y_{t}$, temporal context $y_{t_p},y_{h_p},\hat{y}_{t-1}$.\\
    \textbf{Output:} $\mu_t,\sigma_t$.
    \begin{algorithmic}[1]
    \State Resample $y_t^p,y_h^p$, and $\hat{y}_{t-1}$ using TCR;
    \State Fuse the resampled temporal context using swin transformer encoder;
    \State Apply random masking to current latent:$y_t^S = y_{t}+M$;
    \State Perform cross-attention between the fused context and $y_t^S$ using teacher network;
    \State Cross-attention calculation between fused context and $y_t^S$ using teacher network;
    \State Generate attention map $A$ and entropy map $H$ based on cross-attention output, and perform dependency-weighted processing $\alpha H+ (1-\alpha)A$;
    \State Apply soft top-k selection and remove the mask at selected positions;
    \State Decode the entire latent according to the initially decoded context;
    \State Obtain $\mu_t,\sigma_t$ through linear mapping.
    \end{algorithmic}
    \label{alg:train}
\end{algorithm}

\end{document}